\def\year{2021}\relax
\documentclass[letterpaper]{article} %
\usepackage{aaai21}  %
\usepackage{times}  %
\usepackage{helvet} %
\usepackage{courier}  %
\usepackage[hyphens]{url}  %
\usepackage{graphicx} %
\usepackage{natbib}  %
\usepackage{caption} %
\usepackage{xcolor}
\usepackage{amsfonts}
\usepackage{amsmath}
\usepackage{enumerate,mdwlist}

\title{Extending LIME for Business Process Automation}
\author{Sohini Upadhyay\textsuperscript{\rm 1, \rm 2}, Vatche Isahagian\textsuperscript{\rm 1}, Vinod Muthusamy\textsuperscript{\rm 1}, Yara Rizk\textsuperscript{\rm 1}}

\affiliations {
    \\
    \textsuperscript{\rm 1} IBM Research \\
    \textsuperscript{\rm 2} Harvard University \\
    supadhyay@g.harvard.edu, vatchei@ibm.com,
    vmuthus@us.ibm.com,
    yara.rizk@ibm.com
}

\begin{document}

\maketitle

\begin{abstract}
AI business process applications automate high-stakes business decisions where there is an increasing demand to justify or explain the rationale behind algorithmic decisions. Business process applications have ordering or constraints on tasks and feature values that cause lightweight, model-agnostic, existing explanation methods like LIME to fail. In response, we propose a local explanation framework extending LIME for explaining AI business process applications. Empirical evaluation of our extension underscores the advantage of our approach in the business process setting. 
\end{abstract}

\section{Introduction}
Business processes are an integral part of several industries, including government, insurance, banking and healthcare. Examples of such processes include loan origination, invoice management, automobile insurance claims processing, handling prescription drug orders, and patient case management ~\cite{van2011process}. The business process management (BPM) industry is expected to approach \$16 billion by 2023 \cite{Marketwatch}. Recent advancements in AI represent a great opportunity for infusing AI to predict outcomes~\cite{breuker2016comprehensible}, reduce cost or provide better customer experience~\cite{rao2017sizing}, and recommend decisions~\cite{mannhardt2016decision}. Most recently, deep learning models have been used to make outcome and time-to-complete predictions~\cite{tax2017predictive,evermann2017predicting}.

Unfortunately, very little of these innovations have been applied and adopted by enterprise companies~\cite{daugherty2018human}. One culprit that hinders the adoption of AI business process applications is the ability to explain model decisions. Consider, for example, a mortgage loan application. In the United States, the Federal Trade Commission guidelines dictate that if consumers are denied something of value (i.e, a loan) based on AI, they are entitled to an explanation.
Furthermore, they state that when assigning risk scores to consumers, the key features affecting said scores ought to be disclosed in rank order of importance \cite{ftc}. In the credit scoring domain, this right to an explanation is protected under the Equal Credit Opportunity Act \cite{ecoa}. The European Union guarantees the right to explanation across a broader range of domains under the General Data Protection Regulation (GDPR) \cite{Goodman_2017}. 

The AI community has made significant advancements in the domain of explanation approaches, with explainability tools such as LIME \cite{ribeiro2016should} and SHAP \cite{lundberg2017unified}. Unfortunately, as we will discuss at length, applying these approaches to BPM models directly results in potentially misleading explanations \cite{JanIAAI2020}. 

In this paper, we propose an extension to a popular AI explanation framework, LIME, for business process applications. 
We arrive at this extension by (1) understanding business processes, examining the state of explainable AI, and motivating our choice of base method (LIME), (2) determining the pain points of LIME in the context of business process applications, and (3) developing an extension of LIME and evaluating it in a simulated business process application setting.

\section{Background}
\begin{figure*}[tbp!]
    \centering
    \includegraphics[width=0.8\linewidth]{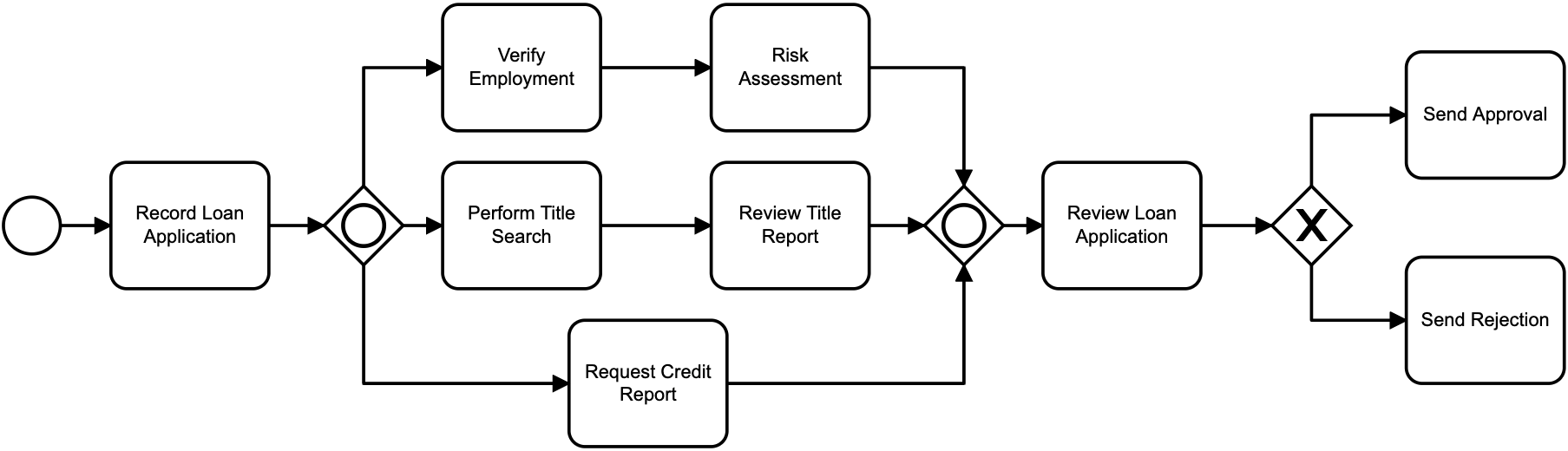}
    \caption{Example of a mortgage loan application process.}
    \label{fig:loan}
\end{figure*}

\subsection{Business Processes}
A business process is a collection of connected tasks that once completely executed delivers a service or product to a client or accomplishes an organizational goal within an enterprise \cite{weske2012bpm}. A mortgage loan application (c.f. Figure \ref{fig:loan}) is an example of a business process where the process flow is the set of linked tasks such as collecting client related data (e.g., verifying employment, requesting credit report, performing a title search, risk assessment, and so on). The goal of the process is to approve or reject a loan application once all the required tasks are fully executed. Business processes are typically modeled using Business Process Model and Notation (BPMN), a graphical notation where activities are denoted by rounded-corner rectangles, and diamonds depict gateways that allow paths to conditionally merge or diverge \cite{grosskopf2009process}.

Business Process Management (BPM) is a multi-disciplinary field that supports the management of business processes with some combination of modeling, automation, execution, control, measurement and optimization. BPM involves business activity flows (workflows), systems, and people such as employees, customers and partners within and beyond the enterprise boundaries. 

\subsection{Explainable AI}
Approaches in explainable AI exhibit various properties \cite{lipton2016mythos}. Some methods are amenable to explanation because they are inherently transparent. In this family of methods, the entire model process or its components are intuitively understandable, like in the case of decision trees or linear models. Other techniques provide post-hoc explanations by extracting information from models that may or may not be inherently transparent themselves. Post-hoc explanations are appealing in deployed applications as they don't require retraining. These techniques feature prominently across the range of explainable AI products and toolkits, including IBM's AI Explainability 360 \cite{AI360}, Microsoft's Azure \cite{azure}, and Google's Explainable AI \cite{google}. Amongst post-hoc approaches, model agnostic methods like LIME and SHAP are particularly flexible as they don't require access to internal model parameters nor knowledge of the model type \cite{ribeiro2016should,lundberg2017unified}. We focus our attention towards LIME as it is the more computationally efficient of the two, with the hopes of adapting and assessing the extension we develop in this paper to both methods in future work. At a high level, LIME constructs explanations by sampling a neighborhood around a data sample and prediction of interest, and training a linear model that mirrors the unknown global model on that neighborhood. The feature coefficients of that linear model are then presented as an explanation of the prediction. As such, on its own, LIME can only provide local explanations. However, methods like MAME and submodular pick can aggregate LIME explanations for multi-level and global model explanations respectively \cite{ramamurthy2020model,ribeiro2016should}. 

\subsection{Explaining BPM Applications}
One major weakness of LIME lies in its local neighborhood construction. LIME generates neighborhood samples by perturbing features independently, even when features are not independent. \cite{slack2019fooling} illustrates that LIME samples can be distributed differently than the true data and exploits this difference for adversarial attacks. \cite{kumar2020problems} notes that this out-of-distribution scenario can be problematic even outside of targeted attack settings. Due to its sampling procedure, LIME samples can take on feature values, or combinations of feature values, that are independently or jointly improbable or even impossible, forcing the local linear model to extrapolate to new parts of the feature space.

This application-agnostic sampling can result in a local linear model that doesn't faithfully reflect the unknown target model that we seek to explain. This is particularly pernicious in the BPM setting as there are often business process rules underpinning feature values and relationships, making LIME  samples more prone to being out-of-distribution. For example, in loan approval settings, features like \textit{Credit} and  \textit{Risk} are inversely correlated; LIME samples could include cases where both are very high or low, scenarios impossible in real life that the target model we are trying to explain may not have seen in training nor handle consistently. In other domains, delays in processing orders are typically correlated with process reworks. 

Another example is with models that predict the path or outcome of a partial execution of a business process~\cite{lstm-bpm2019}. Some of these models, such as those based on LSTMs, learn correlations between the ordering or timing of the activities in the process and the target feature. Perturbations of the timestamp or duration features of these activities can result in activity orderings that violate the process definition or are unlikely to occur. 

This phenomena is not limited to these settings as there are many other settings where such correlations exists. With this in mind, we refer to any prior knowledge about the distribution of BPM data as a business process rule.  
\par

\begin{figure*}[h!]
    \centering
    \includegraphics[width=0.6\linewidth]{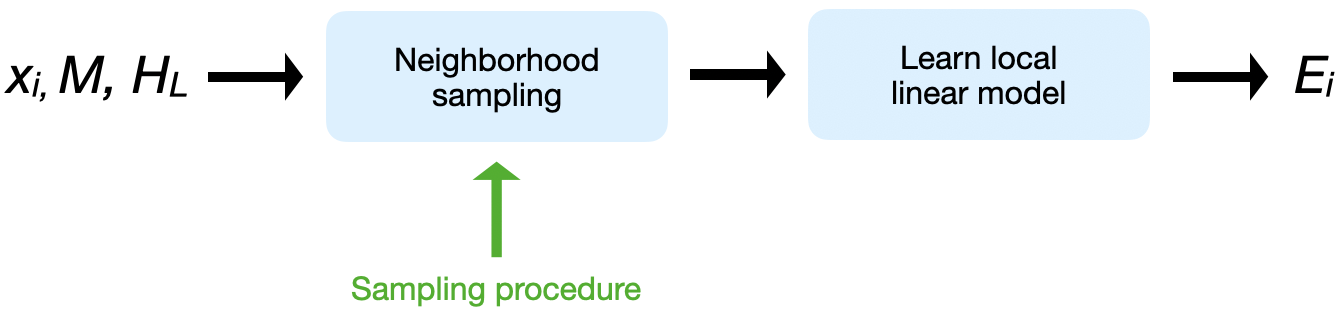}
    \caption{LIME system with extension to specify sampling procedure}
    \label{fig:arch}
\end{figure*}

The full impact of this out-of-distribution problem in the BPM domain is unknown due to the scarcity of case studies of applying LIME to BPM applications. \cite{intuitiaai} creates a custom solution to explaining a credit risk scoring application by employing SHAP, a similar post-hoc explanation method like LIME, along with feature engineering that better encodes knowledge of an important feature. Notice that such knowledge also falls under the scope of business process rules. \cite{JanIAAI2020} hypothesize that the existence of BPM rules could pose a problem to LIME and suggest sampling according to these rules as a direction for solutions. We align this problem with LIME's known out-of-distribution problem and operationalize this suggestion, thereby formalizing an extension to LIME leveraging business process rules.

\section{Methodology}
We extend LIME by modularizing its neighborhood sampling component and exposing it to subject matter experts to define on an application basis according to relevant business process rules. As summarized in Figure \ref{fig:arch}, for a data sample $x_i$, model $M$, and set of LIME hyperparameters $H_L$, LIME samples a random neighborhood around $x_i$, learns a local linear model mirroring the behavior of $M$ on said neighborhood, and returns an explanation of $E_i$ of the output of model $M$ on $x_i$ in terms of the learned local linear model feature coefficients. Our extension operates on the first step, enabling user specification of the sampling procedure. 
\par We illustrate how this can be embodied and how this can mitigate the out-of-distribution problem in a simulation with the following data, business process rule, loan approval model, and explanation components.

\subsection{Data}

\begin{figure}[h!]
    \centering
    \includegraphics[width=\linewidth]{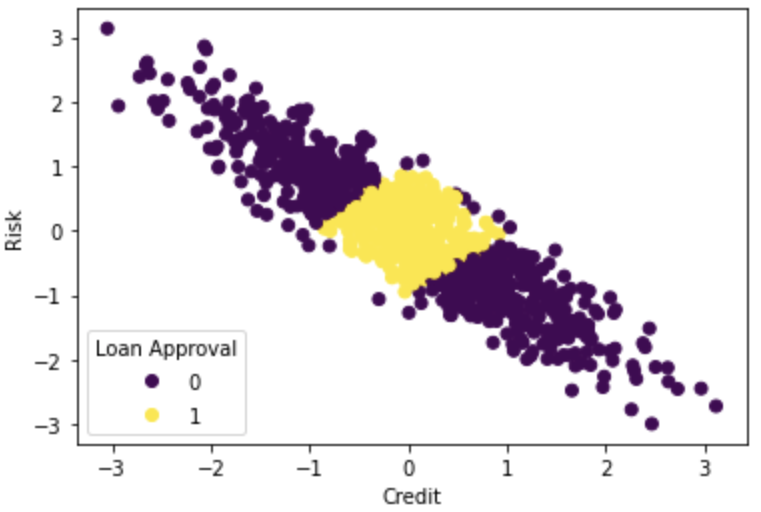}
    \caption{Data distribution}
    \label{fig:a_sim1}
\end{figure}

We simulate a loan approval dataset capturing the scenario where features \textit{Credit} and \textit{Risk} are inversely correlated. We fix $\rho=-0.9$ as the correlation coefficient between \textit{Credit} and \textit{Risk} generate 10,000 samples from a bivariate Gaussian distribution $D$ with $\mu = [0,0]$ and $\Sigma = \rho\mathbb{I}$. Accordingly, each data sample $x_i = [c_i, r_i]$ consists of a \textit{Credit} feature  $c_i$ and \textit{Risk} feature $r_i$. We assign a loan approval label $y_i$ to each data sample $x_i$ with the following rule

\begin{equation}
    y_i = \begin{cases} 
      1 & |c_i+r_i|<1 \text{ and } |c_i-r_i|<1\\
      0 & |c_i+r_i|>=1 \text{ and } |c_i-r_i|>=1
      \end{cases}
\end{equation}

This results in the dataset illustrated in Figure \ref{fig:a_sim1}. 

\subsection{Business process rule}
In this simulation, we take the full relationship between \textit{Credit} and \textit{Rule} to be known to the subject matter experts employing the explanation models. As \textit{Credit} and \textit{Risk} are the only features, we can say the business process rule underpinning this application is that $x_i \sim \mathcal{N}(\mu, \Sigma)$ with $\mu$ and $\Sigma$ as defined above. 

\subsection{Loan approval model}

\begin{figure}[h!]
    \centering
    \includegraphics[width=\linewidth]{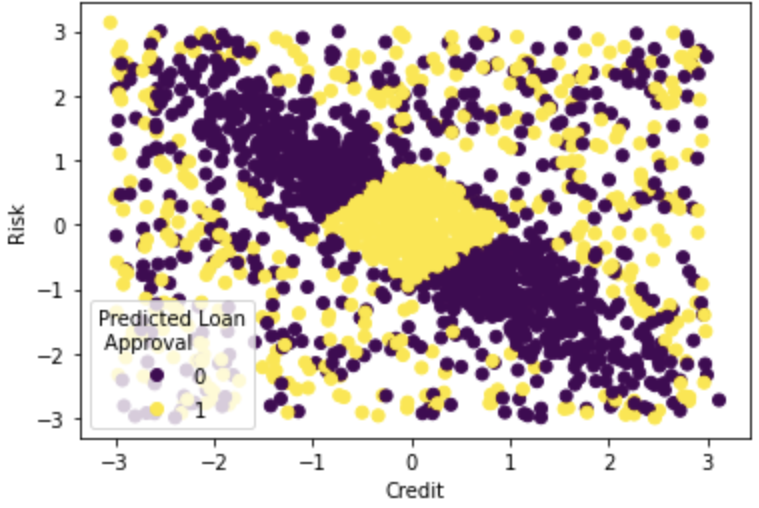}
    \caption{Loan approval predictions of model $M$ on uniform random data points}
    \label{fig:a_sim3}
\end{figure}

\begin{figure*}[h!]
    \centering
    \includegraphics[width=0.8\linewidth]{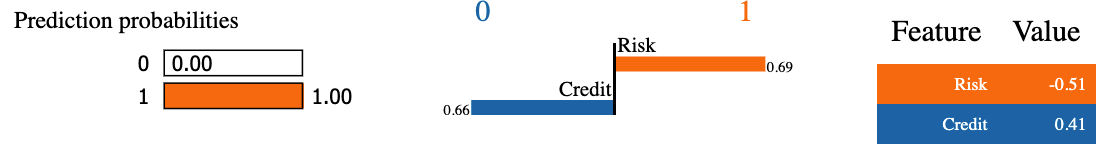}
    \caption{Example LIME explanation}
    \label{fig:lime}
\end{figure*}

One objective of this work is to assess LIME approaches in the presence of the out-of-distribution problem. Therefore, we formulate our loan approval model $M$ such that it is highly accurate on samples that conform to the distribution in Figure \ref{fig:a_sim1} and random on samples that do not conform to the distribution. This is as if a model performs well on real world, process conforming samples, and poorly on impossible, process non-conforming samples. Given the label assignment function $y_i$ as defined by Equation (1), the probability density function $p(x)$ of $D$, and the Bernoulli distribution $Ber(p)$, we define our model as follows:
\begin{equation}
    M(x_i) =  \begin{cases} 
      y_i & p(x_i)>=0.01\\
      Ber(0.5) & p(x_i)<0.01
      \end{cases}
\end{equation}

We illustrate the behavior of model $M$ on a uniform range of data points in Figure \ref{fig:a_sim3}. Observe that there is only a clearly delineated globally nonlinear but locally linear decision boundary for data points that fall within distribution $D$.
As the in-distribution decision boundary follows the label assignment, it is defined in each quadrant as:

\begin{align}
\text{I. } 1-Credit-Risk=0 \label{q1}\\
\text{II. } 1+Credit-Risk=0 \label{q2}\\
\text{III. } 1+Credit+Risk=0 \label{q3}\\
\text{IV. } 1-Credit+Risk=0 \label{q4}
\end{align}

\par While real world BPM models are unlikely to be perfectly accurate on in-distribution, process conforming examples, the accuracy of a model should not significantly affect LIME as the true labels are not used in learning the local linear model. Real world BPM models are also unlikely to behave perfectly randomly on out-of-distribution, process nonconforming samples; they may behave randomly or systematically in a way that may obfuscate the decision boundary near neighboring process conforming samples to a greater or lesser degree. This means that the severity of the out-of-distribution problem may vary on a per-application basis. We choose to study it with this simulation in the extreme case to better understand and improve on worse case behavior of LIME in BPM settings.

\subsection{LIME explanations}

As seen in Figure \ref{fig:arch}, LIME takes the data sample $x_i$ and model $M$ as input, and first samples a neighborhood $N_i$ around $x_i$. The LIME library released by the authors of the original LIME paper defines $N_i$ for continuous valued data differently depending on the choice of certain LIME hyperparameters $\mu_i, \epsilon_i \in H_L$. $\mu_i$ dictates where to center the neighborhood and $\epsilon_i$ determines what type of random noise to use \cite{lime_code}. Choices for $\mu_i$ are centering around the sample, $\mu_x$, or around the training data mean, $\mu_m$. Choices for $\epsilon_i$ are Gaussian noise, $\epsilon_g$, or LHS noise, $\epsilon_l$. Given these hyperparameters, each sample in $N_i$ takes on the form 
\begin{equation}
x_{i}' = \mu_i + \epsilon_i
\label{lime_sampling}
\end{equation}
Our extension allows us to specify new sampling rules based on the business process rules. Then given our business process rule that $x_i \sim \mathcal{N}(\mu, \Sigma)$, we can generate process aware samples directly based on the data distribution such that samples in $N_i$ take on the form 
\begin{equation}
x_{i}' \sim \mathcal{N}(\mu, \Sigma) 
\label{pa_sampling}
\end{equation}
\par 
Once a neighborhood $N_i$ is formed, whether via LIME's standard method as defined in Equation \ref{lime_sampling}, or via process aware sampling, as defined in Equation \ref{pa_sampling}, a local linear model $f_i$ mirroring $M$ on that neighborhood is found via the following optimization:
\begin{equation}
    f_i = \underset{f\in F}{\mathrm{argmin}}\hspace{0.1cm} \mathcal{L}(M, f, \pi_x) + \Omega(f)
\label{fi}
\end{equation}
Here $F$ is the family of linear functions, $\mathcal{L}$ is a weighted loss function, $\pi_x = \pi_x(z)$ is a weight function measuring the proximity between $x=x_i$ and any $z=x_{i}'$, and $\Omega$ is a function penalizing complexity. These parameters are set when specifying $H_L$. Throughout our experiments, we fix $F$ to be the family of logistic regression functions, $\mathcal{L}$ to be weighted square loss, and $\Omega$ to be $L_2$ regression. We use \cite{ribeiro2016should} definition for $\pi_x$, where $\pi_x(z)$ is an exponential kernel of the form 
\begin{equation}
    \pi_x(z) = exp\left(\frac{-D(x,z)^{2}}{\sigma^{2}}\right)
\label{pi}
\end{equation}
for $D,\sigma \in H_L$, where $D$ is a distance function and $\sigma$ is referred to as kernel width. Intuitively, $\sigma$ dictates how large the radius of $N_i$ should be. We use \cite{lime_code} default hyperparameter guidance for $H_L$ unless specified explicitly otherwise. 
\par 

The feature coefficients of $f_i$ then dictate $E_i$. We enclose an example explanation $E_i$ in Figure \ref{fig:lime}. The leftmost column refers to the predicted class probabilities of model $M$ on $x_i$, in this case $1$ for class $1$. The middle column represents the feature coefficients of $f_i$, $-0.66$ and $0.69$ for \textit{Credit} and \textit{Risk}, respectively. The rightmost column reminds us of values of $x_i$, in this example $0.41$ for \textit{Credit} and $-0.51$ for \textit{Risk}.

\section{Evaluation}

\begin{table}[h!] \centering
\begin{tabular}{l|l|l}
 & \multicolumn{2}{c}{$|N_i|=1K$} \\ \cline{2-3} 
 & \multicolumn{1}{c|}{\textit{Credit}} & \multicolumn{1}{c}{\textit{Risk}} \\ \hline
Standard & $0.95\pm 1.01$ & $0.99\pm1.03$ \\ \hline
Process-aware & $0.76 \pm 0.72$ & $0.78\pm 0.80$  
\end{tabular}
\begin{tabular}{l|l|l}
 & \multicolumn{2}{c}{$|N_i|=5K$} \\ \cline{2-3} 
 & \multicolumn{1}{c|}{\textit{Credit}} & \multicolumn{1}{c}{\textit{Risk}} \\ \hline
Standard & $1.46 \pm 3.89$ & $2.61\pm 11.47$ \\ \hline
Process-aware & $0.63\pm 0.54$ & $0.69\pm 0.77$
\end{tabular}
\caption{Coefficient mismatch for neighborhoods with 1000 and 5000 samples, standard and process-aware sampling}
\label{results}
\end{table}

Recall that our model $M$ has a linear decision boundary in each quadrant on in-distribution, process conforming data points. Then, for a test set of our data, we have a ground truth local linear boundary of $M$ defined by these quadrant-wise linear components. We thus hope that LIME explanations result from local linear models that match these ground truth linear components. Accordingly, we measure the accuracy of LIME explanations using the coefficient mismatch metric \cite{rope}. In this setting, coefficient mismatch measures the average absolute differences between the coefficients of $f_i$, the learned local linear model, and the ground truth linear component of $M$ closest to $x_i$. By this definition, coefficient mismatch can take on any value in $[0,\mathbb{R}]$ where the closer it is to 0, the more accurate the explanation is.
 
\par To better illustrate coefficient mismatch, let's calculate it for the explanation in Figure \ref{fig:lime}. From the rightmost column we see that \textit{Credit} is $0.41$ and \textit{Risk} is $-0.51$, placing $x_i$ in the fourth Cartesian quadrant. The ground truth linear component of $M$ in the fourth quadrant is Equation \ref{q4}, where the coefficients of \textit{Credit} and \textit{Risk} are $-1$ and $1$ respectively. As listed in the middle column, in the learned local linear model $f_i$, the coefficient of \textit{Credit} is $-0.66$ in and the coefficient of \textit{Risk} is $0.69$. Thus coefficient mismatch is $0.34$ for \textit{Credit} and $0.31$ for \textit{Risk}. In the experiments below, we average values over multiple trials.

\par We compare explanations from LIME with standard neighborhood sampling to those generated by LIME with process aware neighborhood sampling over 100 trials, across neighborhood sizes of 1000 and 5000 samples, with the former being the standard neighborhood size in \cite{lime_code}. As summarized in Table \ref{results}, we observe that process aware sampling results in consistently lower coefficient mismatch scores for both features, indicating that it yields more accurate explanations. Observe that this trend prevails even when we reduce the neighborhood size, suggesting that process aware sampling requires fewer samples to attain a fixed level of explanation accuracy, or in other words, is more computationally efficient.

\section{Discussion}
Our results suggest that process-aware sampling results in more accurate LIME explanations in BPM settings. As part of our future work, we look to confirm this result on real world BPM data and models. Our major goal is to deliver this capability as part of the IBM's Digital Business Automation suite of products, and making it available in IBM's AI Explainability 360 toolkit. We would expand any explanation user interface to include a textual explanation of the learned local linear model coefficients as the probabilistic interpretation of regression coefficients may not be known to all users.

On the methodological front, we hope to explore more ways to integrate business process knowledge into explanation models. In LIME, BPM rules could also be leveraged in $D$, the distance function hyperparameter weighting neighborhood samples, particularly in the case of categorical data. If business process rules imply that a group of categories are more similar to each other than others, then that could inform the distance to neighborhood samples with those features more than a standard measure of whether the categories are identical or not. We also hope to explore whether our extension can be applied to similar explainable AI methods like SHAP where neighborhood sampling is also employed and the same out-of-distribution problem is observed \cite{lundberg2017unified,slack2019fooling,kumar2020problems}. Recall that \cite{intuitiaai} needed to perform additional feature engineering to encode knowledge about an important feature before using SHAP. If we take such knowledge to be a business process rule, incorporating process aware sampling into SHAP could perhaps resolve this in a systematic way with case-study agnostic guidelines. 

\bibliographystyle{aaai}
\bibliography{references}

\end{document}